\pdfoutput=1

\documentclass[11pt]{article}

\usepackage[final]{acl}

\usepackage{times}
\usepackage{latexsym}

\usepackage[T1]{fontenc}

\usepackage[utf8]{inputenc}

\usepackage{microtype}

\usepackage{inconsolata}

\usepackage{graphicx}

\usepackage{booktabs}
\usepackage{multirow}
\usepackage{amsmath}
\usepackage{arydshln}
\usepackage{adjustbox}
\usepackage{color,xcolor,colortbl}
\usepackage[framemethod=TikZ]{mdframed}
\usepackage{pgfplots}
\usepgfplotslibrary{fillbetween}
\usetikzlibrary{patterns}
\usepackage{subcaption}

\pgfplotsset{compat=1.18}

\makeatletter
\def\adl@drawiv#1#2#3{%
        \hskip.5\tabcolsep
        \xleaders#3{#2.5\@tempdimb #1{1}#2.5\@tempdimb}%
                #2\z@ plus1fil minus1fil\relax
        \hskip.5\tabcolsep}
\newcommand{\cdashlinelr}[1]{%
  \noalign{\vskip\aboverulesep
           \global\let\@dashdrawstore\adl@draw
           \global\let\adl@draw\adl@drawiv}
  \cdashline{#1}
  \noalign{\global\let\adl@draw\@dashdrawstore
           \vskip\belowrulesep}}
\makeatother

\newcommand{\quotes}[1]{``#1''}
\renewcommand{\vec}[1]{\mathbf{#1}}

\definecolor{lightblue}{HTML}{5DA5DA}
\definecolor{lightorange}{HTML}{FAA43A}
\definecolor{lightgreen}{HTML}{60BD68}
\definecolor{lightpurple}{HTML}{B276B2}
\definecolor{brown}{HTML}{D95F0E}
\definecolor{lightgrey}{HTML}{BDBDBD}
\usepackage{fancyhdr}

\fancypagestyle{firstpage}{
  \fancyhf{} 
  \fancyfoot[C]{\thepage} 
  \fancyfoot[L]{\scriptsize Accepted at ACL 2025. Official version: \href{https://doi.org/10.18653/v1/2025.acl-long.41}{doi.org/10.18653/v1/2025.acl-long.41}} 
}

\title{LACA: Improving Cross-lingual Aspect-Based Sentiment Analysis with LLM Data Augmentation}

\author{
 Jakub \v{S}m\'{i}d\textsuperscript{*, $\dagger$} \and
 Pavel P\v{r}ib\'{a}\v{n}\textsuperscript{*} \and
 Pavel Kr\'{a}l\textsuperscript{*, $\dagger$}
 \vspace{8pt}
\\
 \textsuperscript{*}Department of Computer Science and Engineering\\
 \textsuperscript{$\dagger$}NTIS -- New Technologies for the Information Society\\
 University of West Bohemia in Pilsen, Faculty of Applied Sciences \\
         Univerzitní 2732/8, 301 00 Pilsen, Czech Republic \\
    \texttt{\{jaksmid,pribanp,pkral\}@kiv.zcu.cz}\\
         \url{https://nlp.kiv.zcu.cz}
}

\begin{document}
\maketitle
\thispagestyle{firstpage} 

\begin{abstract}
Cross-lingual aspect-based sentiment analysis (ABSA) involves detailed sentiment analysis in a target language by transferring knowledge from a source language with available annotated data. Most existing methods depend heavily on often unreliable translation tools to bridge the language gap. In this paper, we propose a new approach that leverages a large language model (LLM) to generate high-quality pseudo-labelled data in the target language without the need for translation tools. First, the framework trains an ABSA model to obtain predictions for unlabelled target language data. Next, LLM is prompted to generate natural sentences that better represent these noisy predictions than the original text. The ABSA model is then further fine-tuned on the resulting pseudo-labelled dataset. We demonstrate the effectiveness of this method across six languages and five backbone models, surpassing previous state-of-the-art translation-based approaches. The proposed framework also supports generative models, and we show that fine-tuned LLMs outperform smaller multilingual models. 
\end{abstract}

\section{Introduction}
Aspect-based sentiment analysis (ABSA) is a natural language processing (NLP) task that identifies sentiments linked to specific aspects within a sentence~\citep{liu2010sentiment}, often used to evaluate products or services. For example, in the sentence \textit{\quotes{Great tea but terrible service}}, the aspect terms are \textit{\quotes{tea}} with positive sentiment and \textit{\quotes{service}} with negative sentiment. E2E-ABSA aims to extract aspect terms and their associated sentiment polarities together. The wide-ranging applications of ABSA have garnered substantial interest in recent years~\citep{zhang2022survey}. Nevertheless, research has primarily focused on English, leaving other languages largely unexplored due to the lack of annotated data. However, manual labelling is and time-consuming and costly, especially for low-resource languages, making cross-lingual ABSA a valuable research area. This work explores zero-shot cross-language ABSA, which leverages annotated source language data to transfer knowledge to target languages without labelled data.

Early cross-lingual ABSA research used machine translation with alignment algorithms~\citep{lambert-2015-aspect, zhou2015clopinionminer} and cross-lingual word embeddings~\citep{barnes-etal-2016-exploring, akhtar-etal-2018-solving} to transfer knowledge between languages. Multilingual pre-trained language models (mPLMs) like mBERT~\citep{devlin-etal-2019-bert} and XLM-R~\citep{conneau-etal-2020-unsupervised} have become standard in capturing cross-lingual syntactic and semantic patterns, forming the basis for recent advancements~\citep{zhang-etal-2021-cross, lin2023clxabsa, LinXABSA}, though challenges persist in zero-shot transfer due to language-specific aspect terms, slang, and abbreviations in real-world texts~\citep{li2020unsupervised}.

Cross-lingual ABSA faces challenges, especially in zero-shot settings, as models fine-tuned on source language data can struggle with language-specific aspect terms and informal language~\citep{SMID2025103073}. Additionally, many low-resource languages are underrepresented in mPLMs' pre-training corpora~\citep{conneau-etal-2020-unsupervised}, and manual annotation for ABSA is time and resource-intensive. While translation-based methods offer a solution, they often introduce noise by misaligning aspect terms, leading to partial or missing terms in the target language~\citep{li2020unsupervised}. This misalignment disrupts the model’s ability to correctly identify aspect terms in the target language, reducing cross-lingual ABSA accuracy.

Recent advances in large language models (LLMs) open new possibilities for cross-lingual ABSA. LLM-based data augmentation, which generates diverse examples in the target language without translation, is a promising yet underexplored alternative to machine translation for cross-lingual ABSA. Similarly, fine-tuning LLMs for cross-lingual ABSA remains largely unexplored, despite their success in English~\citep{smid-etal-2024-llama}. This paper addresses these gaps by proposing a novel LLM-based data augmentation approach leveraging unlabelled target language data as an alternative to machine translation and exploring LLM fine-tuning for cross-lingual ABSA.

To this end, we propose the \textbf{L}LM \textbf{A}ugmented \textbf{C}ross-lingual \textbf{A}BSA (\textsc{LACA}) framework, which leverages unlabelled target language data to improve cross-lingual ABSA performance. The framework begins by fine-tuning an ABSA model on labelled source language data $\mathcal{D}_\mathcal{S}$. The model then predicts a label $\boldsymbol{\hat{y}}^\mathcal{T}$ for each unlabelled sentence $\boldsymbol{x}^\mathcal{T}$ from the target language dataset
$\mathcal{D}_\mathcal{T}$. To reduce prediction noise caused by language differences, we prompt an LLM with each predicted label $\boldsymbol{\hat{y}}^\mathcal{T}$ to generate a corresponding target language sentence~$\boldsymbol{\hat{x}}^\mathcal{T}$.
This step ensures the generated data better aligns with the predicted labels than the original unlabelled data, thereby reducing prediction noise. Next, we pair each generated target language sentence  $\boldsymbol{\hat{x}}^\mathcal{T}$ with its corresponding predicted label $\boldsymbol{\hat{y}}^\mathcal{T}$ to form a new pseudo-labelled dataset $\mathcal{D}_\mathcal{G}$. Finally, this dataset is combined with the source language dataset $\mathcal{D}_\mathcal{S}$ to train a final model. Our proposed approach provides a powerful alternative to traditional translation-based methods, fully utilizes unlabelled target language data, and effectively addresses the language gap issue by transforming noisy predictions into more accurate text-label pairs. By generating target language sentences that explicitly align with predicted labels, our framework reduces inconsistencies caused by direct cross-lingual prediction, ensuring better adaptation to linguistic nuances. \textsc{LACA} boosts cross-lingual ABSA performance, achieving 1.50\% and 2.62\% average improvements over previous state-of-the-art methods across two models.

Our key contributions are: 1) We introduce a novel \textsc{LACA} framework, which enhances cross-lingual ABSA by generating high-quality pseudo-labelled target language data using LLMs, effectively avoiding the language gap problems by generating coherent natural sentences given noisy predicted labels. 2) We demonstrate the effectiveness and robustness of the proposed approach across six languages and five backbone models, achieving new state-of-the-art results. 3) We show that the proposed framework is adaptable to generative models, highlighting its versatility. 4) We find that fine-tuned LLMs outperform smaller multilingual models, being the first to underscore the advantages of LLMs for cross-lingual ABSA.

\section{Related Work}

Early cross-lingual ABSA research primarily targets simple tasks, focusing on a single sentiment element. Common approaches to cross-lingual transfer include machine translation~\citep{lambert-2015-aspect, klinger-cimiano-2015-instance, zhou2015clopinionminer} and cross-lingual word embeddings~\citep{wang2018transition, jebbara-cimiano-2019-zero, akhtar-etal-2018-solving, barnes-etal-2016-exploring}.

Recent research mainly targets E2E-ABSA and utilizes mPLMs such as mBERT~\citep{devlin-etal-2019-bert} and XLM-R~\citep{conneau-etal-2020-unsupervised}, often in combination with machine translation. Techniques to further improve performance include parameter warm-up~\citep{li2020unsupervised}, alignment-free label projection with distillation on unlabelled data~\citep{zhang-etal-2021-cross}, contrastive learning for semantic alignment~\citep{lin2023clxabsa}, and dynamic weighted loss to address class imbalances~\citep{LinXABSA}.

\begin{figure*}[ht!]
    \centering
    \includegraphics[width=0.9\linewidth]{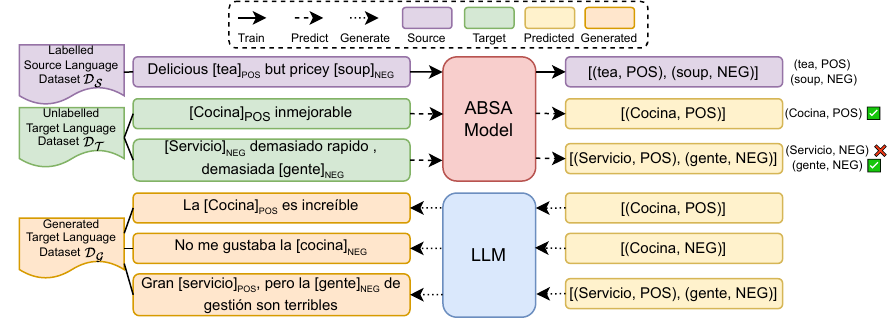}
    \caption{The proposed \textsc{LACA} framework integrates fine-tuning and predictions with the ABSA model and pseudo-labelled data generated by an LLM. Square brackets denote gold aspect terms and their polarities. Gold labels for the target language (Spanish) are included for illustration purposes only. The generated dataset is later merged with the labelled source dataset for the final ABSA model training.}
    \label{fig:overview}
\end{figure*}

LLMs tend to underperform compared to smaller models fine-tuned for ABSA~\citep{gou-etal-2023-mvp,zhang2023sentiment}, though fine-tuned LLaMA models achieve state-of-the-art results in monolingual ABSA~\citep{smid-etal-2024-llama,icaart25}. Several studies leverage LLMs for data augmentation~\citep{li-etal-2022-controllable, moller-etal-2024-parrot, ding-etal-2024-data}, including for English ABSA~\citep{zhong2024iterativedatagenerationlarge}.

\section{Methodology}
This section describes our \textbf{L}LM \textbf{A}ugmented \textbf{C}ross-lingual \textbf{A}BSA (\textsc{LACA}) framework. Figure~\ref{fig:overview} illustrates its two main stages: first, fine-tuning the ABSA model on labelled source language data to make predictions on unlabelled target language data (top part); second, using an LLM to generate high quality data to match these predictions to create pseudo-labelled target language dataset (bottom part), which is then used for further training of the ABSA model.

\subsection{Problem Formulation}

ABSA involves analyzing a sentence $\boldsymbol{x} = (x_i)_{i=1}^n$ containing $n$ tokens. This task can be framed as a sequence labelling problem. The model predicts a sequence of labels $\boldsymbol{y}=(y_i)_{i=1}^n$, where $y_i \in \mathcal{Y}$ is selected from the label space $\mathcal{Y} = \mathtt{\{B, I\}\text{-}\{POS, NEG, NEU\}}\cup \{\mathtt{O}\}$. These labels capture the boundaries and sentiment of aspect terms in the sentence, such as $y_i=\mathtt{B\text{-}NEU}$ for the beginning of a neutral aspect term.


Alternatively, the ABSA task can be formulated as a text generation problem, where the model predicts a set of sentiment tuples $\boldsymbol{y}=\{(a_i, p_i)\}_{i=1}^T$, where each tuple consists of an aspect term $a_i$ and its corresponding sentiment polarity $p_i$. The number of tuples $T$ depends on the input sentence.

In cross-lingual settings, the goal is to predict a label $\boldsymbol{y}^\mathcal{T}$ for a sentence $\boldsymbol{x}^\mathcal{T}$ in the target language~$\mathcal{T}$, using only sentence-label pairs $(\boldsymbol{x}^\mathcal{S}, \boldsymbol{y}^\mathcal{S})$ from the source language $\mathcal{S}$ in the dataset $\mathcal{D}_\mathcal{S}$, without access to labelled data from the target language. However, unlabelled target language sentences from the dataset $\mathcal{D}_\mathcal{T} = \{\boldsymbol{x}_i\}_{i=1}^{|\mathcal{D}_\mathcal{T}|}$ can assist the task.

\subsection{ABSA Models}
\label{sub:model}
We use pre-trained multilingual models as the backbone of our ABSA model, denoting the parameters as $\vec{\Theta}$, which includes task-specific parameters $\vec{W}$ and $\vec{b}$, all fine-tuned during training.

For sequence labelling, we employ encoder-based models that convert the input sequence ${\boldsymbol{x} = (x_i)_{i=1}^n}$ into hidden vectors $\vec{h} = (\vec{h}_i)_{i=1}^n$. A linear classification layer produces token-level predictions from hidden vectors using BIO tagging for aspect boundaries and sentiment polarities. The label distribution for each token $x_i$ is computed as
\begin{equation}
    P_\vec{\Theta}(y_i|x_i)={\operatorname{softmax}}(\vec{W}\vec{h}_i + \vec{b}).
\end{equation}
We minimize the cross-entropy loss $\mathcal{L}$ between the predicted and true labels as
\begin{equation}
    \small
    \mathcal{L} = \frac{1}{|\mathcal{D}|}\sum_{(\boldsymbol{x},\boldsymbol{y})\in \mathcal{D}}\left[-\frac{1}{n}\sum_{i=1}^ny_i\log P_\vec{\Theta}(y_i|x_i)\right].
\end{equation}

We also explore the ABSA task as a text-generation problem, using sequence-to-sequence (encoder-decoder) and decoder-only models. In sequence-to-sequence models, the encoder processes the input sequence $x$ into a contextualized representation $\vec{e}$. The decoder generates the output sequence $\boldsymbol{y}$ token by token, with each token $y_i$ predicted based on the previous tokens $y_{1}^{i-1}$ and the encoded input $\vec{e}$. We format the output as \quotes{{\sffamily[A]} $a$ {\sffamily[P]} $p$}, where $a$ represents the aspect term and $p$ its corresponding sentiment polarity, concatenating multiple outputs with {\sffamily[;]}. During fine-tuning, we minimize the cross-entropy as 
\begin{equation}
    \small
    \mathcal{L} = \frac{1}{|\mathcal{D}|}\sum_{(\boldsymbol{x},\boldsymbol{y})\in \mathcal{D}}\left[-\frac{1}{n}\sum_{i=1}^n\log P_{\vec{\Theta}}(y_i|\vec{e},y_1^{i-1})\right].
\end{equation}
Decoder-only models function similarly, except they generate tokens solely based on previously generated tokens, without relying on encoded input sequences.

\subsection{Pseudo-Labelled Data Generation}
While the ABSA model can make predictions directly in the target language, research has shown that pseudo-labelled target language data improves cross-lingual ABSA performance~\citep{zhang-etal-2021-cross}. A straightforward method for generating pseudo-labels without machine translation is to pair each target language sentence $\boldsymbol{x}^\mathcal{T}$ with its corresponding model prediction $\boldsymbol{\hat{y}}^\mathcal{T}$. However, this self-training approach can be hindered by noise in the predictions. To address this, we propose employing LLMs for data augmentation, generating sentences that align better with the predicted labels.

Specifically, we input the predicted label $\boldsymbol{\hat{y}}^\mathcal{T}$ into the LLM, prompting it to generate a sentence~$\boldsymbol{\hat{x}}^\mathcal{T}$ that matches the label. As a result, the LLM generates a pseudo-labelled dataset $\mathcal{D}_{\mathcal{G}}$ consisting of $(\boldsymbol{\hat{x}}_i^\mathcal{T}, \boldsymbol{\hat{y}}_i^\mathcal{T})$ pairs. As discussed, the gap between the source and target languages introduces noise into the ABSA model’s predictions on unlabelled target language data. Instead of refining the ABSA model, our LLM-based augmentation creates more reliable pseudo-labelled training samples, where each sentence $\boldsymbol{\hat{x}}^\mathcal{T}$ accurately reflects the predicted label $\boldsymbol{\hat{y}}^\mathcal{T}$, thereby minimizing the impact of the noise in the predictions.

Pseudo-labels are crucial for exposing the model to language-specific elements like slang and aspect terms in the target language, which pre-training alone cannot fully address. They help bridge the gap between source and target languages by encouraging the model to learn and adapt to the target language's nuances.

We improve the LLM’s understanding by providing ten few-shot examples from the source language training data, rotating these examples randomly to ensure the diversity of the output. Due to the limited number of sentiment polarities and the natural diversity of aspect terms, this random selection is sufficient to produce varied and representative examples. Additionally, we can modify the input examples as needed to address imbalances in the source language training set. For instance, if certain sentiment polarities are underrepresented, we can create new inputs that reflect different sentiment polarities while preserving the aspect term. This strategy helps generate more diverse examples and also aids in mitigating class imbalances within the dataset.

Unlike machine translation methods that translate source language data directly into the target language – often resulting in semantically similar examples – our LLM-based approach is designed to generate a more diverse set of target language examples. While translation methods yield two linguistically distinct datasets, the underlying semantics remain largely unchanged. In contrast, the proposed LLM augmentation introduces a wide range of semantically distinct examples, enhancing the model's generalization and robustness by exposing it to a broader spectrum of meanings. Furthermore, we can adjust the LLM inputs to address label imbalances, generating data for less frequent sentiment polarities as needed.

\subsection{Training}
To ensure the quality of the generated dataset $\mathcal{D}_{\mathcal{G}}$, it should meet several key criteria: generated sentences should accurately reflect all sentiment elements in the tuples, include only the specified sentiment elements, and be in the target language.

\begin{figure}[ht!]
    \centering
    \includegraphics[width=\linewidth]{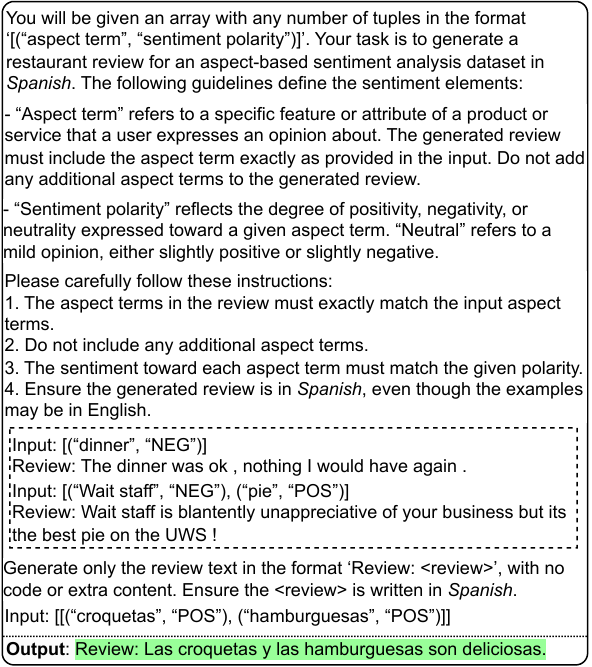}
    \caption{LLM prompt illustration for review generation, with two few-shot demonstrations in the dashed box and the expected output in the green box. The example uses \textit{Spanish} but is adaptable to other languages.}
    \label{fig:promptgenerate}
\end{figure}

To achieve the quality of $\mathcal{D}_{\mathcal{G}}$, we pre-process the predicted labels $\boldsymbol{\hat{y}}^\mathcal{T}$ to guarantee that at least one sentiment element is present. We also craft the generation prompt to specify that the text must be in the target language and not introduce additional sentiment elements, as shown in Figure~\ref{fig:promptgenerate}. After generating pairs $(\boldsymbol{\hat{x}}^\mathcal{T}, \boldsymbol{\hat{y}}^\mathcal{T})$, we post-process them by filtering out instances where $\boldsymbol{\hat{x}}^\mathcal{T}$ lacks aspect terms from $\boldsymbol{\hat{y}}^\mathcal{T}$. We also discard pairs where the ABSA model's prediction on $\boldsymbol{\hat{x}}^\mathcal{T}$ differs from~$\boldsymbol{\hat{y}}^\mathcal{T}$.

Finally, we combine the source language dataset $\mathcal{D}_{\mathcal{S}}$ with the generated dataset $\mathcal{D}_{\mathcal{G}}$ to form the final training set, continuing the training of the same model as described in Section~\ref{sub:model}.

\section{Experimental Setup}
We conduct experiments on the E2E-ABSA task.

\subsection{Dataset}
We evaluate the proposed framework on the SemEval-2016 dataset~\citep{pontiki-etal-2016-semeval}, which includes real user restaurant reviews in English (en), Spanish (es), French (fr), Dutch (nl), Russian (ru), and Turkish (tr). We use the data splits provided by \citet{zhang-etal-2021-cross} for a fair comparison. Table~\ref{tab:data} shows the dataset statistics.
\begin{table}[ht!]
    \centering
    \begin{adjustbox}{width=\linewidth}   
        \begin{tabular}{@{}llrrrrrr@{}}
        \toprule
                               &     & \textbf{En} & \textbf{Es} & \textbf{Fr} & \textbf{Nl} & \textbf{Ru} & \textbf{Tr} \\ \midrule
        \multirow{2}{*}{Train} & No. sentences & 1,600       & 1,656       & 1,332       & 1,378       & 2,924 & 986      \\
                               & No. aspects & 1,377       & 1,500       & 1,294       & 956         & 2,439 & 1,083       \\ \cdashlinelr{1-8}
        \multirow{2}{*}{Dev}   & No. sentences & 400         & 414         & 322        & 344         & 731 & 246        \\
                               & No. aspects & 365         & 353         & 345         & 274         & 629   & 271      \\\cdashlinelr{1-8}
        \multirow{2}{*}{Test}  & No. sentences & 676         & 881         & 668         & 575         & 1,209   & 144    \\
                               & No. aspects & 612         & 713         & 649         & 373         & 945   & 148      \\ \bottomrule 
        \end{tabular}
    \end{adjustbox}
    \caption{Data statistics for each language.}
    \label{tab:data}
\end{table}

In all experiments, we use the source language validation set for model selection to ensure true unsupervised settings~\citep{jebbara-cimiano-2019-zero}.

\subsection{Implementation Details}
We employ base mBERT~\citep{devlin-etal-2019-bert} and XLM-R~\citep{conneau-etal-2020-unsupervised} for the encoder models based on related work~\citep{li2020unsupervised, zhang-etal-2021-cross, lin2023clxabsa, LinXABSA}, base mT5~\citep{xue-etal-2021-mt5} for sequence-to-sequence models, and Orca~2~13B~\citep{mitra2023orca2teachingsmall} and LLaMA~3.1 8B~\citep{dubey2024llama3herdmodels} for decoder-only models.

For the LLMs generating the pseudo-labelled examples, we employ Orca~2~13B and LLaMA~3.1~8B and 70B. To diversify the dataset and reduce sentiment imbalance, we modify 20\% of over-represented positive sentiment examples by generating new instances, with a 60\% chance of neutral and 40\% of negative sentiment. Appendix~\ref{sec:experiments} presents the detailed experimental details.

\subsection{Evaluation Metrics}
We employ micro-F1 as the evaluation metric, consistent with related work~\citep{zhang-etal-2021-cross, lin2023clxabsa, LinXABSA}, where a prediction is deemed correct only if both its boundary and sentiment polarity are accurate. We report average F1 scores across five runs with different random seeds.

\subsection{Compared Methods}
We compare our approach against the \textsc{Zero-shot} method, which fine-tunes the model using only labelled source language data, a strong baseline for cross-lingual tasks~\citep{conneau-etal-2020-unsupervised, wu-dredze-2019-beto}, and several translation-based approaches. \textsc{Translation-TA} employs the \textit{Translate-then-Align} paradigm~\citep{li2020unsupervised} for fine-tuning using translated data, while \textsc{Bilingual-TA} combines this translated data with the original source data. \textsc{ACS}~\citep{zhang-etal-2021-cross} uses an alignment-free projection method and aspect code-switching to interchange aspect terms between languages. \textsc{ACS-Distill} enhances this by applying distillation on unlabelled target language data. \textsc{CL-XABSA}~\citep{lin2023clxabsa} incorporates contrastive learning at both the sentiment (\textsc{SL}) and token levels (\textsc{TL}). \textsc{Equi-XABSA}~\citep{LinXABSA} employs a dynamically weighted loss to address class imbalances and anti-decoupling to enhance semantic information utilization.

\section{Results}
\begin{table*}[ht!]
    \centering
    \begin{adjustbox}{width=0.9\linewidth}
        \begin{tabular}{@{}lcccccccccc@{}}
            \toprule
            \multirow{2}{*}{\textbf{Method}} & \multicolumn{5}{c}{\textbf{mBERT}}    & \multicolumn{5}{c}{\textbf{XLM-R}}    \\  \cmidrule(lr){2-6} \cmidrule(lr){7-11}
                                             & Es    & Fr    & Nl    & Ru    & Avg   & Es    & Fr    & Nl    & Ru    & Avg   \\ \midrule
            \textsc{Supervised} {\footnotesize \citep{zhang-etal-2021-cross}}                     & 67.88 & 61.80 & 56.80 & 58.87 & 61.34 & 71.93 & 67.44 & 64.28 & 64.93 & 67.15 \\ \cdashlinelr{1-11}
            \textsc{Zero-shot}                         & 56.90       & 45.80       & 45.97       & 34.06       & 45.68 & 67.48       & 58.87       & 58.95       & 56.10       & 60.35 \\
            \textsc{Translation-TA} {\footnotesize \citep{li2020unsupervised}}                        & 50.71 & 40.76 & 47.13 & 41.67 & 45.08 & 58.10 & 47.00 & 56.19 & 50.34 & 52.91 \\
            \textsc{Bilingual-TA} {\footnotesize \citep{li2020unsupervised}}                           & 51.23 & 41.00 & 49.72 & 43.67 & 46.41 & 61.87 & 49.34 & 58.64 & 52.89 & 55.69 \\
            \textsc{ACS} {\footnotesize \citep{zhang-etal-2021-cross}}                       & 59.99 & 49.65 & 51.19 & 52.09 & 53.23 & 67.32 & 59.39 & 62.83 & 60.81 & 62.59 \\
            \textsc{ACS-Distill} {\footnotesize \citep{zhang-etal-2021-cross}}               & 62.91 & 52.25 & 53.40 & \textbf{54.58} & 55.79 & 69.24 & 59.90 & 63.74 & 62.02 & 63.73 \\
            \textsc{CL-XABSA (TL)} {\footnotesize \citep{lin2023clxabsa}}                 & 60.64 & 48.53 & 50.96 & 50.77 & 52.73 & 64.85 & 58.10 & 59.75 & 58.84 & 60.39 \\
            \textsc{CL-XABSA (SL)} {\footnotesize \citep{lin2023clxabsa}}                 & 61.62 & 49.50 & 50.64 & 50.65 & 53.10 & 64.63 & 59.47 & 59.40 & 61.13 & 61.16 \\
            \textsc{Equi-XABSA} {\footnotesize \citep{LinXABSA}}                      & 63.08 & 50.08 & 51.85 & 52.59 & 54.40 & 69.56 & 60.68 & 61.31 & 62.34 & 63.47 \\ \cdashlinelr{1-11}
            \textsc{\textbf{LACA\textsubscript{LLaMA\textsubscript{70}}}}                           & \textbf{65.23} & \textbf{54.90} & \underline{55.29} & 53.72 & \textbf{57.29} & \textbf{71.89} & \textbf{64.97} & \underline{65.35} & \underline{63.20} & \textbf{66.35} \\
            \textsc{\textbf{LACA\textsubscript{Orca\textsubscript{13}}}}                             & \underline{64.80} & \underline{54.21} & \textbf{55.41} & \underline{53.86} & \underline{57.07} & \underline{71.61} & \underline{64.25} &\textbf{65.41} & \textbf{63.46} & \underline{66.18} \\ 
            \textsc{\textbf{LACA\textsubscript{LLaMA\textsubscript{8}}}} & 64.33 & 	53.74 & 	54.56 &	52.36 &	56.25 &	71.17 &	63.81 &	64.29 & 	61.46 & 65.18 \\
            
            \bottomrule
        \end{tabular}
    \end{adjustbox}
    \caption{Average F1 scores over five runs with different random seeds for cross-lingual E2E-ABSA using English as the source language, compared with supervised (monolingual) results in the \quotes{\textsc{Supervised}} row and cross-lingual results from other studies. The best scores are highlighted in \textbf{bold}, and the second-best scores are \underline{underlined}.}
    \label{tab:main_results}
\end{table*}

Table~\ref{tab:main_results} presents the cross-lingual ABSA results using mBERT and XLM-R as backbone models. Key observations include: 
\\
\hspace*{5pt} 
1) XLM-R is a strong baseline in \textsc{Zero-shot} settings, while mBERT underperforms. 
\\
\hspace*{5pt} 
2) \textsc{Translation-TA} and \textsc{Bilingual-TA} perform similarly or worse than \textsc{Zero-shot}.
\\
\hspace*{5pt} 
3) The leading translation-based approaches are \textsc{ACS-Distill}, which uses distillation on unlabelled target data, and \textsc{Equi-XABSA}, which addresses class imbalances. 
\\
\hspace*{5pt} 
4) Our framework with LLaMA~3.1~8B (\textbf{\textsc{LACA\textsubscript{LLaMA\textsubscript{8}}}}) surpasses the best results of translation-based methods by around 0.5\% with mBERT and over 1\% with XLM-R on average.
\\
\hspace*{5pt} 
5) The proposed method using Orca~2~13B (\textbf{\textsc{LACA\textsubscript{Orca\textsubscript{13}}}}) outperforms prior methods in all languages except Russian with mBERT, showing a 1.28\% improvement over the best translation-based methods with mBERT and 2.45\% with XLM-R, while enhancing \textsc{Zero-shot} by 11.39\% and 5.83\% on average, respectively. It sets new state-of-the-art results for Dutch with both models and Russian with XLM-R. Despite being English-centric\footnote{The official paper~\citep{mitra2023orca2teachingsmall} does not specify supported languages, but since Orca~2 is built on LLaMA~2~\citep{touvron2023llama2openfoundation}, it probably primarily targets English.}, Orca~2~13B for \textsc{LACA} outperforms the smaller multilingual LLaMA~3.1~8B and nearly matches the larger LLaMA~3.1~70B, surpassing it on Russian and Dutch, languages not officially supported by LLaMA~3.1. This ability to rival the larger multilingual model may stem from Orca~2's advanced reasoning capabilities. Additionally, Orca~2 tends to generate shorter reviews, potentially reducing errors such as introducing aspect terms not present in predicted labels, which can harm the ABSA model performance.
\\
\hspace*{5pt} 
6) \textsc{LACA} with LLaMA~3.1~70B (\textbf{\textsc{LACA\textsubscript{LLaMA\textsubscript{70}}}}) achieves new state-of-the-art results with mBERT and XLM-R in Spanish, French, and on average. It surpasses the previous best methods by 1.50\% with mBERT and 2.62\% with XLM-R while improving the \textsc{Zero-shot} baseline by 11.61\% with mBERT and 6\% with XLM-R. The 70B version of LLaMA~3.1 outperforms the 8B version by more than 1\% on average, demonstrating that larger models offer better performance but at the expense of slower inference and higher memory usage.
\\
\hspace*{5pt} 
7) Notably, XLM-R with \textsc{LACA\textsubscript{Orca\textsubscript{13}}} and \textsc{LACA\textsubscript{LLaMA\textsubscript{70}}} matches the performance of supervised settings in Spanish and exceeds it in Dutch, while being less than 1\% below average performance across all languages. Crucially, our approach achieves this without the need for external translation tools.
\\
\hspace*{5pt}
8) Spanish performs best as the target language, likely due to its similarity to English, which leads to better-aligned embeddings in pre-trained models. The LLMs also tend to generate higher-quality examples in Spanish due to their stronger representation in that language. 
\\
\hspace*{5pt} 
9) Though strong, our performance in Russian is slightly lower than in other languages, likely due to its greater dissimilarity to English and the lack of official LLaMA~3.1 support, which may reduce the quality of generated examples. In contrast, Dutch benefits from its similarity to supported languages like English and German, despite not being officially supported by LLaMA~3.1.

\begin{table}[ht!]
    \centering
    \begin{adjustbox}{width=0.9\linewidth}
        \begin{tabular}{@{}lccccc@{}}
        \toprule
                  & \textbf{Es}          & \textbf{Fr}          & \textbf{Nl}          & \textbf{Ru}          & \textbf{Avg}         \\ \midrule
        mBERT     & 56.90                & 45.80                & 45.97                & 34.06                & 45.68                \\
        \phantom{0}+\textsc{\textbf{LACA\textsubscript{LLaMA\textsubscript{70}}}}     & \underline{65.23}          & \underline{54.90}          & 55.29          & 53.72          & \underline{57.29}          \\
        \phantom{0}+\textsc{\textbf{LACA\textsubscript{Orca\textsubscript{13}}}}     & 64.80                & 54.21                & \underline{55.41}                & \underline{53.86}                & 57.07                \\
        \phantom{0}+\textsc{\textbf{LACA\textsubscript{LLaMA\textsubscript{8}}}}     & 64.33 & 	53.74 & 	54.56 &	52.36 &	56.25                \\
        \cdashlinelr{1-6}
        XLM-R     & 67.48                & 58.87                & 58.95                & 56.10                & 60.35                \\
        \phantom{0}+\textsc{\textbf{LACA\textsubscript{LLaMA\textsubscript{70}}}}     & \underline{71.89}          & \underline{64.97}          & 65.35          & 63.20 & \underline{66.35}          \\
        \phantom{0}+\textsc{\textbf{LACA\textsubscript{Orca\textsubscript{13}}}}     & 71.61                & 64.25                & \underline{65.41}                & \textbf{\underline{63.46}}                & 66.18 \\
        \phantom{0}+\textsc{\textbf{LACA\textsubscript{LLaMA\textsubscript{8}}}}     & 71.17 &	63.81 &	64.29 & 	61.46 & 65.18
        \\ \cdashlinelr{1-6}
        mT5       & 66.85                & 58.12                & 58.47                & 55.65                & 59.77                \\
        \phantom{0}+\textsc{\textbf{LACA\textsubscript{LLaMA\textsubscript{70}}}}     & \underline{72.03}          & \underline{63.92}          & 64.95          & 62.71          & 65.90          \\
        \phantom{0}+\textsc{\textbf{LACA\textsubscript{Orca\textsubscript{13}}}}     & 71.56                & 63.49                & \underline{65.70}                & \underline{62.92}                & \underline{65.92} \\
        \phantom{0}+\textsc{\textbf{LACA\textsubscript{LLaMA\textsubscript{8}}}}     & 70.56                & 62.99               & 63.70                & 60.92                & 64.54
        \\ \cdashlinelr{1-6}
        LLaMA 3.1 & 69.24                & 66.02                & 64.74                & 55.14                & 63.79                \\
        \phantom{0}+\textsc{\textbf{LACA\textsubscript{LLaMA\textsubscript{70}}}}     & 73.74 & \underline{\textbf{70.73}} & \underline{68.04} & 62.49          & \underline{68.75} \\
        \phantom{0}+\textsc{\textbf{LACA\textsubscript{Orca\textsubscript{13}}}}     & \underline{73.75}                & 70.39                & 67.95                & \underline{62.68}                & 68.69                \\
        \phantom{0}+\textsc{\textbf{LACA\textsubscript{LLaMA\textsubscript{8}}}}     & 73.20                & 69.89                & 67.95                & 60.92                & 67.81 \\
        \cdashlinelr{1-6}
        Orca 2 & 69.35                & 65.93                & 64.85                & 55.21                & 63.84                \\
        \phantom{0}+\textsc{\textbf{LACA\textsubscript{LLaMA\textsubscript{70}}}}     & \underline{\textbf{74.27}} & \underline{70.13} & 68.25 & 62.38          & \underline{\textbf{68.76}} \\
        \phantom{0}+\textsc{\textbf{LACA\textsubscript{Orca\textsubscript{13}}}}     & 73.80                & 69.89                & \textbf{\underline{68.47}}                & \underline{62.49}                & 68.66                \\
        \phantom{0}+\textsc{\textbf{LACA\textsubscript{LLaMA\textsubscript{8}}}}     & 73.21                & 69.30                & 67.48                & 60.09                & 67.52 \\
        
        \bottomrule
        \end{tabular}
    \end{adjustbox}
    \caption{Results for different fine-tuned models in zero-shot settings and with \textsc{LACA}. The best results for each model are \underline{underlined}, and the best overall are in \textbf{bold}.}
    \label{tab:res_all_models}
\end{table}

Table~\ref{tab:res_all_models} shows the results of our approach with five different backbone models. The mT5 model performs similarly to XLM-R, indicating that the sequence-to-sequence approach  can effectively serve as an alternative to sequence labelling methods. LLaMA~3.1 and Orca~2 consistently yield the best results across languages, except for Russian, likely due to the lower support for Russian in these LLMs. The \textsc{LACA} framework improves the performance of LLaMA~3.1 and Orca~2 by nearly 5\% compared to the zero-shot approach. On average, \textsc{LACA} with LLMs as backbone models outperforms XLM-R by more than 2\%, highlighting the potential of fine-tuned LLMs for cross-lingual ABSA tasks. However, the larger parameter count of LLaMA~3.1 (about 30 times larger than XLM-R) and Orca~2 (about 50 times larger than XLM-R) results in slower inference times and higher GPU memory requirements, presenting a trade-off when opting for LLMs over smaller models.

\subsection{Results for Turkish}
Table~\ref{tab:res_tr} presents the results for Turkish as the target language. Most prior research~\citep{zhang-etal-2021-cross, lin2023clxabsa, LinXABSA} has excluded Turkish due to the very small test set, containing fewer than 150 examples. Despite this limitation, the proposed \textsc{LACA} framework demonstrates significant improvements for both mBERT and XLM-R models compared to \textsc{Zero-shot} and translation-based approaches. These results highlight the framework's adaptability and effectiveness for languages outside the Indo-European family.

\begin{table}[ht!]
    \centering
    \begin{adjustbox}{width=0.9\linewidth}
        \begin{tabular}{@{}lcc@{}}
            \toprule
            \textbf{Method}                                               & \textbf{mBERT}       & \textbf{XLM-R}       \\ \midrule
            \textsc{Supervised}                                           & 47.74  & 60.93 \\ \cdashlinelr{1-3}
            \textsc{Zero-shot}                                            & 27.04                & 46.53                \\
            \textsc{Translation-TA} {\footnotesize \citep{li2020unsupervised}}                                      & 22.04                & 40.24                \\
            \textsc{Bilingual-TA} {\footnotesize \citep{li2020unsupervised}}                                      & 22.64                & 41.44                \\ \cdashlinelr{1-3}
            \textsc{\textbf{LACA\textsubscript{LLaMA\textsubscript{70}}}} & \underline{31.98}          & \underline{50.02}          \\
            \textsc{\textbf{LACA\textsubscript{Orca\textsubscript{13}}}}  & \textbf{33.16}       & \textbf{51.15}       \\
            \textsc{\textbf{LACA\textsubscript{LLaMA\textsubscript{8}}}}  & 31.13                & 49.71                \\ \bottomrule
        \end{tabular}
    \end{adjustbox}
    \caption{Results for Turkish as the target language with English as the source language. The best results for each model are in \textbf{bold}; the second best are \underline{underlined}.}
    \label{tab:res_tr}
\end{table}

\subsection{Additional Results}
Appendix~\ref{sec:add_res} shows additional results with smaller LLMs (LLaMA~3.2~1B and 3B) and different source-target language combinations, further showcasing the effectiveness of the proposed method.

\subsection{Ablation Study}
Table~\ref{tab:ablation} shows an ablation study of \textsc{LACA}, highlighting the impact of its key components.

\begin{table}[ht!]
    \centering
    \begin{adjustbox}{width=0.95\linewidth}
        \begin{tabular}{@{}lccccc@{}}
            \toprule
                                                           & \textbf{Es} & \textbf{Fr} & \textbf{Nl} & \textbf{Ru} & \textbf{Avg} \\ \midrule
            \multicolumn{6}{c}{\textbf{mBERT}}                                   \\
            \textbf{\textsc{LACA\textsubscript{LLaMA\textsubscript{70}}}}                                    & \textbf{65.23}                  & \textbf{54.90}                  & \underline{55.29}                  & \underline{53.72}                  & \textbf{57.29}                   \\
            \phantom{0}– w/o extra example creation               & 64.73                     & \underline{54.32}                     & 53.76                     & 51.50                     & 56.08                      \\
            \phantom{0}– w/o dynamic few-shot                          & 63.87                           & 52.46                           & 52.29                           & 49.01                           & 54.41                            \\
            LLaMA 70B  text \& label gen. & 57.32                           & 43.62                           & 46.08                           & 39.15                           & 46.54                            \\
             LLaMA 70B  translation gen. & 63.02                           & 52.15                           & 52.68                           & 51.54                           & 54.85                            \\
            \textbf{\textsc{LACA\textsubscript{Orca\textsubscript{13}}}}                                     & \underline{64.80}                           & 54.21                           & \textbf{55.41}                           & \textbf{53.86}                           & \underline{57.07}                            \\
            \phantom{0}– w/o extra example creation               & 63.72                           & 53.28                           & 53.74                           & 51.98                           & 55.68                            \\
            \phantom{0}– w/o dynamic few-shot                           & 62.77                           & 52.08                           & 52.97                           & 49.46                           & 54.32                            \\
            Orca 13B  text \& label gen.  & 56.87                           & 42.12                           & 46.31                           & 39.94                           & 46.31                            \\
            Orca 13B  translation gen. & 62.83                           & 52.17                           & 52.99                           & 51.81                           & 54.95                            \\
            \textbf{\textsc{LACA\textsubscript{LLaMA\textsubscript{8}}}}                                     & 64.33 & 	53.74 & 	54.56 &	52.36 &	56.25                            \\
            \phantom{0}– w/o extra example creation               & 63.72                           & 53.28                           & 53.74                           & 52.60                           & 55.84                            \\
            \phantom{0}– w/o dynamic few-shot                           & 62.77                           & 52.08                           & 52.97                           & 49.46                           & 54.32                            \\
            LLaMA 8B text \& label gen.  & 56.17                           & 40.23                           & 44.64                           & 35.27                           & 44.07                            \\
            LLaMA 8B  translation gen. & 62.43                           & 51.91                           & 52.29                           & 51.12                           & 54.44                            \\
            Continue (MLM pre-train)                                  & 57.26                           & 46.02                           & 47.21                           & 37.56                           & 47.01                            \\
            Self-training                                     & 53.84                           & 38.97                           & 36.77                           & 25.04                           & 38.66                            \\ \cdashlinelr{1-6}
            \multicolumn{6}{c}{\textbf{XLM-R}}                                  \\
            \textbf{\textsc{LACA\textsubscript{LLaMA\textsubscript{70}}}}                                    & \textbf{71.89}                  & \textbf{64.97}                  & \underline{65.35}                  & \underline{63.20}                  & \textbf{66.35}                   \\
            \phantom{0}– w/o extra example creation              & 71.35                     & 64.18                     & 64.25                     & 62.07                     & 65.46                      \\
            \phantom{0}– w/o dynamic few-shot                          & 70.12                           & 62.98                           & 63.34                           & 59.52                           & 63.99                            \\
            LLaMA 70B  text \& label gen. & 66.50                           & 59.12                           & 58.29                           & 57.01                           & 60.23                            \\
            LLaMA 70B  translation gen. & 70.09                           & 61.69                           & 62.88                           & 61.20                           & 63.97                            \\
            \textbf{\textsc{LACA\textsubscript{Orca\textsubscript{13}}}}                                     & \underline{71.61}                           & \underline{64.25}                           & \textbf{65.41}                           & \textbf{63.46}                           & \underline{66.18}                            \\
            \phantom{0}– w/o extra example creation               & 70.48                           & 63.09                           & 64.48                           & 62.63                           & 65.17                            \\
            \phantom{0}– w/o dynamic few-shot                           & 69.81                           & 63.34                           & 62.27                           & 60.36                           & 63.63                            \\
            Orca 13B text \& label gen.  & 64.28                           & 58.12                           & 58.99                           & 58.28                           & 59.92                            \\
            Orca 13B  translation gen. & 69.69                           & 61.21                           & 62.84                           & 61.45                           & 63.80                            \\

            \textbf{\textsc{LACA\textsubscript{LLaMA\textsubscript{8}}}}                                     & 71.17 &	63.81 &	64.29 & 	61.46 & 65.18                           \\
            \phantom{0}– w/o extra example creation               & 70.48                           & 63.09                           & 63.48                           & 60.63                           & 64.42                            \\
            \phantom{0}– w/o dynamic few-shot                           & 69.81                           & 62.34                           & 62.27                           & 59.11                           & 63.38                            \\
            LLaMA 8B text \& label gen.  & 64.28                           & 58.12                           & 58.99                           & 54.28                           & 58.92                            \\
            LLaMA 8B  translation gen. & 69.71                           & 61.03                           & 61.91                           & 60.97                           & 63.41                            \\
             Continue (MLM pre-train)                                  & 68.95                           & 59.12                           & 58.94                           & 58.74                           & 61.44                            \\
            Self-training                                     & 65.67                           & 53.63                           & 57.57                           & 51.48                           & 57.09                            \\ \bottomrule
        \end{tabular}
    \end{adjustbox}
    \caption{Ablation study of the \textsc{LACA} framework, with the best results in \textbf{bold} and the second best \underline{underlined}.}
    \label{tab:ablation}
\end{table}

\paragraph{Effect of additional examples creation}
To investigate the effectiveness of creating additional examples by replacing sentiment polarity, we remove this step and denote it as \quotes{w/o extra example creation}. The results indicate a small improvement (0.5–1.2\% on average) when additional examples are included, suggesting further gains might be possible with more example creation. In preliminary experiments, increasing the sentiment polarity modification ratio from 20\% to 50\% did not improve performance but significantly increased generation time. Researchers should carefully consider the trade-offs between computational cost, practicality, and potential performance gains when modifying sentiment combinations.

\paragraph{Effect of few-shot examples switching}
We also evaluate the impact of maintaining static few-shot examples instead of switching them for each generated sample (\quotes{w/o dynamic few-shot}). Results show a clear performance drop of about 2\% without dynamic examples, confirming that variety in few-shot samples improves generation quality.

\paragraph{Effect of label generation}
To examine the impact of utilizing predicted labels, we replace the pseudo-labelling process with LLMs prompted to generate both text and labels rather than generating text based on provided labels (\quotes{text \& label gen.}). This approach, which does not leverage unlabelled target data, leads to a significant performance drop of around 11\% for mBERT and 6\% for XLM-R. Several factors contribute to this decline. First, the generated labels often have incorrect formats, such as producing \texttt{B-NEUT} instead of the correct \texttt{B-NEU}, leading to discarded examples. Second, while the ABSA model in \textsc{LACA} provides diverse aspect terms for the LLM to generate text, prompting the LLM to generate both text and labels leads to repetitive, single-word aspects, reducing accuracy and diversity. Additionally, the LLM sometimes assigns incorrect sentiments or mislabels aspects, compounding noise in predictions. Research indicates that LLaMA-based models tend to underperform in ABSA in zero-shot and few-shot scenarios~\citep{smid-etal-2024-llama}, suggesting that they are ill-suited for generating coherent examples that include both text and labels.

\paragraph{Effect of prediction-based label generation}
To assess the importance of the initial training and prediction phase, we replaced it with directly translating aspect terms from the English dataset using the Google Translate API and generating pseudo-labelled data with LLMs (\quotes{translation gen.}) from the translations. This approach performed approximately 2\% worse than our method. We attribute this to the prediction phase's ability to identify language-specific aspect terms in the unlabelled target dataset, which may lack direct equivalents in the source dataset. By generating sentences based on these unique terms, our method produces more diverse, target-language-specific data, capturing nuances that simple translation overlooks and enhancing the quality of pseudo-labelled data.

\paragraph{Effect of unlabelled data utilization}
Additional (continued) pre-training on domain-specific corpora has proven effective for leveraging unlabelled data~\citep{xu-etal-2019-bert}. We compare this approach -- labelled as \quotes{Continue} -- with our method to analyze the use of unlabelled data for cross-lingual ABSA. In this setup, we replace the LLM-based generation with masked language model (MLM) pre-training on unlabelled target language data, following the original pre-training objectives of mBERT and XLM-R. Results show that continued pre-training performs 10\% worse with mBERT and 5\% worse with XLM-R than our method. This performance gap is likely attributed to the substantial amount of data required for effective continued pre-training, which is often impractical in cross-lingual ABSA scenarios.

\begin{table}[ht!]
    \centering
    \begin{adjustbox}{width=\linewidth}
        \begin{tabular}{@{}lll@{}}
        \toprule
        \textbf{Spanish Dataset Sentence}                                   & \textbf{Prediction}         & \textbf{\textsc{LACA} Generation}                                 \\ \midrule
        el [servicio]\textsubscript{\texttt{POS}} impecable.                                                 & (servicio, \texttt{POS})             &  \begin{tabular}[c]{@{}l@{}}El [servicio]\textsubscript{\texttt{POS}} fue\\ excelente, rápido y\\ agradable.\end{tabular} \\ 
        {\small (\textit{impeccable service.})} &  & {\small \textit{\begin{tabular}[c]{@{}l@{}}(The service was excellent,\\ fast and friendly.)\end{tabular}}} \\
        \cdashlinelr{1-3}
        \begin{tabular}[c]{@{}l@{}}Magnifíca [atención]\textsubscript{POS},\\ buena [carta de vinos]\textsubscript{\texttt{POS}},\\ muy buen [paella]\textsubscript{\texttt{POS}}.\end{tabular}  & \begin{tabular}[c]{@{}l@{}}(carta, \texttt{POS}),\\ (paella, \texttt{POS})\end{tabular} & \begin{tabular}[c]{@{}l@{}}Buen [carta]\textsubscript{\texttt{POS}},\\{} [paella]\textsubscript{POS} deliciosa.\end{tabular}              \\ 
        {\small \textit{\begin{tabular}[c]{@{}l@{}}(Great service, good wine\\ list, very good paella.)\end{tabular}}} & &  {\small (\textit{Good menu, delicious paella.})}
        \\\cdashlinelr{1-3}
        \begin{tabular}[c]{@{}l@{}}El [servicio]\textsubscript{\texttt{NEU}} era\\ eficiente pero no\\ especialmente amable.\end{tabular}                & (servicio, \texttt{NEG})             & El [servicio]\textsubscript{\texttt{NEG}} era terrible                        \\
        {\small \textit{\begin{tabular}[c]{@{}l@{}}(The service was efficient\\ but not particularly friendly.)\end{tabular}}} & &   {\small (\textit{The service was terrible})}
        \\ 
        \bottomrule
        \end{tabular}
    \end{adjustbox}
    \caption{Examples of data generation for Spanish, with gold aspect terms marked with square brackets.}
    \label{tab:generated}
\end{table}

\paragraph{Effect of pseudo-labelled data generation}
Finally, we replace the proposed method with self-training, fine-tuning the model on pseudo-labelled data generated directly by the ABSA model. This approach leads to significant performance drops (up to 20\% for mBERT and 9\% for XLM-R), primarily due to the noisy predictions in zero-shot settings. Our method, which employs LLM-based generation, successfully mitigates this issue by reducing noise in pseudo-labelled data. We manually reviewed several generated samples across different languages, with a few examples in Spanish presented in Table~\ref{tab:generated}. The gold data for the target language is provided solely for investigation purposes and is not available during training. The second example is missing one aspect term and has one incomplete aspect term, while the third example has incorrectly assigned sentiment polarity. Nevertheless, the LLM can generate accurate sentences that effectively describe the predictions, even if they do not match the original input. These instances illustrate how this stage can address noisy predictions and produce high-quality target language data.

\subsection{Analysis of the Generated Samples}

We analyzed 50 randomly generated examples from the LLaMA~3.1 model with 70B parameters for each target language. To streamline the process, we focused on one model and a limited number of samples, as this analysis is time-intensive. Since we are not native speakers of any of the target languages, we utilized the Google Translate API\footnote{\url{ https://translate.google.com/}} to translate the generated examples (except for English) and input aspect terms into English, acknowledging that this might introduce some noise. Notably, the authors have prior experience annotating datasets for ABSA, which enhances their understanding of the task. For all target languages, English served as the source language. However, for English itself, we used Spanish as the source language.

From the reviewed examples, none were missing the requested aspect term (verified in the original language before translation), except for Turkish, where one instance contained a slightly modified version of the requested term rather than an exact match. We focused on two potential error types:
\begin{enumerate}
    \item \textbf{Introduction of new aspect terms:} Instances where the model generated additional aspect terms that were not requested.
    \item \textbf{Incorrect sentiment polarity:} Cases where the sentiment polarity of the generated text did not align with the expected polarity.
\end{enumerate}

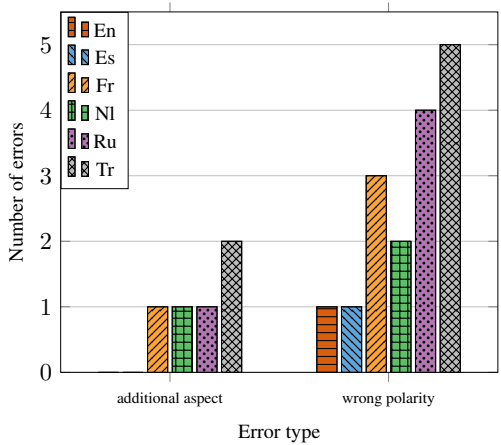
\begin{figure}[ht!]
    \centering
    \begin{adjustbox}{width=0.85\linewidth}
        \begin{tikzpicture}
            \begin{axis}[
                ybar,
                bar width=9pt,
                xtick={0,1},
                ytick={0,1,2,3,4,5},
                xticklabels={\scriptsize \strut additional aspect,\scriptsize \strut wrong polarity},
                ymin=0,
                ymax=5.5,
                xmin=-0.5,
                xmax=1.5,
                ymajorgrids=true,
                ylabel={\footnotesize Number of errors},
                xlabel={\footnotesize Error type},
                legend style={at={(0,1)}, anchor=north west, font=\footnotesize},
                ]
                \addplot[black,fill=brown,postaction={pattern=horizontal lines}] coordinates {(0,0) (1,1)};
                \addplot[black,fill=lightblue,postaction={pattern=north west lines}] coordinates {(0,0) (1,1)};
                \addplot[black,fill=lightorange,postaction={pattern=north east lines}] coordinates {(0,1) (1,3)};
                \addplot[black,fill=lightgreen,postaction={pattern=grid}] coordinates {(0,1) (1,2)};
                \addplot[black,fill=lightpurple,postaction={pattern=crosshatch dots}] coordinates {(0,1) (1,4)};
                \addplot[black,fill=lightgrey,postaction={pattern=crosshatch}] coordinates {(0,2) (1,5)};
                \legend{En,Es,Fr,Nl,Ru,Tr}
            \end{axis}
        \end{tikzpicture}
    \end{adjustbox}
    \caption{Number of error types in 50 samples generated by LLaMA~3.1~70B for different languages.}
    \label{fig:error_generated}
\end{figure}

Figure~\ref{fig:error_generated} summarizes the results of this analysis. New aspect term errors were minimal across all languages, while errors related to incorrect sentiment polarity occurred slightly more often but remained rare overall. Most polarity errors involved the neutral sentiment class, where neutral polarity is expected to indicate slightly positive or slightly negative sentiment. However, some samples that should have been positive or negative were misclassified as neutral, and vice versa. Since the neutral sentiment class accounts for only about 5\% of the samples in each test set, these errors have a negligible impact on the overall performance. The highest number of errors occurred in Turkish, followed by Russian – both languages that are not officially supported by the model.

During the analysis, we have noticed that the model do not tend to produce similar sentences for same sentiment elements and similar aspect terms, likely due to the use of sampling and different few-shot examples for each generated sample.

\subsection{Further Analysis}
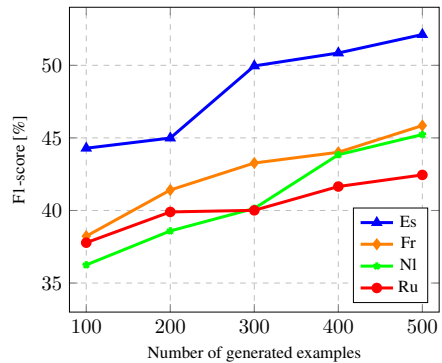
\begin{figure}[ht!]
    \centering
    \begin{adjustbox}{width=0.75\linewidth}
        \begin{tikzpicture}
            \begin{axis}[
                xlabel={\small{Number of generated examples}},
                ylabel={\small{F1-score [\%]}},
                xtick={100, 200, 300, 400, 500},
                xtick=data,
                ymin=33, ymax=54,
                xmin=80, xmax=520,
                ytick={35,40,45,50},
                legend pos=south east,
                ymajorgrids=true,
                xmajorgrids=true,
                grid style=dashed,
                legend style={font=\footnotesize},
                ]

                \addplot[
                    color=blue,
                    mark=triangle*,
                    ultra thick,
                ] 
                coordinates {
                    (100,44.29)
                    (200,44.99)
                    (300,49.96)
                    (400,50.85)
                    (500,52.12)
                };
                \addlegendentry{Es}

                \addplot[
                    color=orange,
                    mark=diamond*,
                    ultra thick,
                ] 
                coordinates {
                    (100,38.23)
                    (200,41.41)
                    (300,43.27)
                    (400,44.01)
                    (500,45.85)
                };
                \addlegendentry{Fr}

                \addplot[
                    color=green,
                    mark=star,
                    ultra thick,
                ] 
                coordinates {
                    (100,36.25)
                    (200,38.59)
                    (300,40.14)
                    (400,43.84)
                    (500,45.23)
                };
                \addlegendentry{Nl}

                \addplot[
                    color=red,
                    mark=*,
                    ultra thick,
                ] 
                coordinates {
                    (100,37.78)
                    (200,39.90)
                    (300,40.01)
                    (400,41.65)
                    (500,42.45)
                };
                \addlegendentry{Ru}

            \end{axis}
        \end{tikzpicture}
	\end{adjustbox}
	\caption{Impact of the number of target language samples generated by our method on XLM-R performance.}
	\label{fig:few}
\end{figure}

Figure~\ref{fig:few} illustrates the results using XLM-R with varying numbers of generated samples. By excluding source training data and relying solely on generated samples for final training, we observe a clear trend: performance improves as the number of generated samples increases, highlighting their effectiveness in enhancing cross-lingual capabilities.

\subsection{Error Analysis}
Appendix~\ref{sec:error_analysis} provides an error analysis, offering additional insights into potential improvements and identifying limitations.

\section{Conclusion}
In this paper, we introduce the \textsc{LACA} framework to enhance cross-lingual ABSA. The proposed approach utilizes a large language model to generate high-quality pseudo-labelled data for the target language based on the predictions provided by the ABSA model. We establish new state-of-the-art results, surpassing translation-based methods, and demonstrate the effectiveness of the proposed framework across six languages and five backbone models. Additionally, we show that sequence-to-sequence approaches, supported by our framework, can serve as a viable alternative to traditional sequence labelling methods. Furthermore, we demonstrate that fine-tuned LLMs consistently outperform smaller multilingual models.

\section*{Limitations}
Despite achieving state-of-the-art performance in cross-lingual ABSA, the proposed framework has some limitations. First, while the experiments confirmed its effectiveness for cross-lingual ABSA, it could be extended to tasks like named entity recognition. Second, the performance of our method improves with larger LLMs, but this also increases training time and demands more computational resources, although it does not affect inference. Smaller LLMs can perform significantly worse than larger ones, especially for unsupported languages. Additionally, performance may be influenced by the target language support of the employed LLM, as unsupported languages may result in lower-quality pseudo-labelled data. This issue can be mitigated by selecting an LLM that explicitly supports the target language, making model choice a crucial factor in achieving strong performance. Another potential issue is that the models may struggle to generate neutral polarity reliably, which could be problematic for datasets where neutral sentiment is more prevalent. Next, budget constraints prevented evaluation with closed-source LLMs. Finally, limited annotated datasets in various languages restrict our evaluation to the restaurant domain.

\section*{Ethics Statement}
We conduct experiments on widely used datasets from previous scientific studies, ensuring a fair and transparent analysis of results. Our work is carried out ethically, with no harm to individuals. However, models used in this study may exhibit unintended biases related to race or gender due to the large pre-training corpus sourced from the Internet.

\section*{Acknowledgements}
The work of Jakub \v{S}m\'{i}d has been supported by the Grant No. SGS-2025-022 -- New Data Processing Methods in Current Areas of Computer Science.
The work of the other authors has been supported by the project R\&D of Technologies for Advanced Digitalization in the Pilsen Metropolitan Area (DigiTech) No. CZ.02.01.01/00/23\_021/0008436. 
Computational resources were provided by the e-INFRA CZ project (ID:90254), supported by the Ministry of Education, Youth and Sports of the Czech Republic.

\bibliography{bibliography}

\appendix

\section{Experiments Details}
\label{sec:experiments}

For all experiments, we use models from the HuggingFace Transformers library\footnote{\url{https://github.com/huggingface/transformers}}~\citep{wolf-etal-2020-transformers}, the AdamW optimizer~\citep{loshchilov2019decoupledweightdecayregularization}, a batch size of 16, and a single NVIDIA L40 GPU with 48 GB memory.

For encoder-based models, we use base versions of mBERT~\citep{devlin-etal-2019-bert} and XLM-R~\citep{conneau-etal-2020-unsupervised}, following prior works~\citep{li2020unsupervised, zhang-etal-2021-cross, lin2023clxabsa, LinXABSA}. The learning rates are set to 5e-5 for mBERT and 2e-5 for XLM-R, with optimal epochs searched within \{10, 15, 20, 25, 30\}.

For sequence-to-sequence models, we use base mT5~\citep{xue-etal-2021-mt5} with a learning rate of 3e-4, epochs searched within \{15, 20, 25\}, employing greedy search as the decoding algorithm.

\begin{figure}[ht!]
    \centering
    \includegraphics[width=\linewidth]{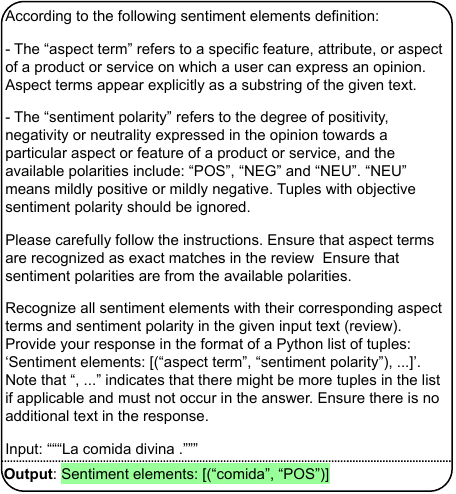}
    \caption{Illustration of the classification LLM prompt, including the expected output in the green box.}
    \label{fig:prompt}
\end{figure}

For decoder-only models, we fine-tune the 8B version of LLaMA~3.1~\citep{dubey2024llama3herdmodels} and the 13B version of Orca~2~\citep{mitra2023orca2teachingsmall} using QLoRA~\cite{dettmers2023qlora} with 4-bit NormalFloat quantization. Following recommendations, we use a constant learning rate of 2e-4 and apply LoRA adapters~\cite{hu2021lora} on all linear transformer block layers, with LoRA parameters $r=64$ and $\alpha=16$. We fine-tune the model for up to 5 epochs with the greedy search for decoding. Figure~\ref{fig:prompt} shows the prompt for fine-tuning.

We employ 4-bit quantized 70B and 8B versions of LLaMA~3.1 and the 13B version of Orca~2 as LLMs for generating pseudo-labelled data. For additional analysis, we also employ 1B and 3B versions of LLaMA~3.2. We use top-$p$ sampling~\citep{holtzman2020curiouscaseneuraltext} with $p=0.8$ and a temperature of $0.8$ to encourage more diverse outputs. When generating new input tuples, we specifically target over-represented positive sentiment examples, modifying 20\% by generating new instances, assigning neutral sentiment with a 60\% chance and negative sentiment with a 40\% chance. This strategy diversifies the dataset and partly addresses sentiment distribution imbalances.

\section{Additional Results}
\label{sec:add_res}
This section provides additional results using smaller LLMs for generation and for different source–target language combinations.

\subsection{Result with Smaller LLMs}
Table~\ref{tab:res_small} presents additional results with smaller LLMs, specifically, the 1B and 3B versions of LLaMA~3.2 (\textbf{\textsc{LACA\textsubscript{LLaMA\textsubscript{1}}}} and \textbf{\textsc{LACA\textsubscript{LLaMA\textsubscript{3}}}}), compared to the main results with larger models. While smaller models still improve over \textsc{Zero-shot} results in all cases, they exhibit performance drops compared to larger LLMs.

\begin{table}[ht!]
    \centering
    \begin{adjustbox}{width=\linewidth}
    \begin{tabular}{@{}lcccccr@{}}
    \toprule
    \textbf{Method}                                               & \textbf{Es}       & \textbf{Fr}       & \textbf{Nl}       & \textbf{Ru}       & \textbf{Tr}       & \multicolumn{1}{c}{\textbf{Avg}} \\ \midrule
    \multicolumn{7}{c}{\textbf{mBERT}}                                                                                                                                                                   \\
    \textsc{Zero-shot}                                            & 56.90             & 45.80             & 45.97             & 34.06             & 27.04             & 41.95                            \\
    \textbf{\textsc{LACA\textsubscript{LLaMA\textsubscript{70}}}} & \textbf{65.23}    & \textbf{54.90}    & \underline{55.29} & \underline{53.72}             & \underline{31.98} & \underline{52.22}                \\
    \textbf{\textsc{LACA\textsubscript{Orca\textsubscript{13}}}}  & \underline{64.80} & \underline{54.21} & \textbf{55.41}    & \textbf{53.86} & \textbf{33.16}    & \textbf{52.29}                   \\
    \textbf{\textsc{LACA\textsubscript{LLaMA\textsubscript{8}}}}  & 64.33             & 53.74             & 54.56             & 52.36             & 31.13             & 51.22                            \\
    \textbf{\textsc{LACA\textsubscript{LLaMA\textsubscript{3}}}}  & 59.15             & 48.83             & 49.94             & 49.99             & 27.59             & 47.10                            \\
    \textbf{\textsc{LACA\textsubscript{LLaMA\textsubscript{1}}}}  & 58.37             & 47.02             & 48.29             & 45.52             & 27.09             & 45.26                            \\ \cdashlinelr{1-7}
    \multicolumn{7}{c}{\textbf{XLM-R}}                                                                                                                                                                   \\
    \textsc{Zero-shot}                                            & 67.48             & 58.87             & 58.95             & 56.10             & 46.53             & 57.59                            \\
    \textbf{\textsc{LACA\textsubscript{LLaMA\textsubscript{70}}}} & \textbf{71.89}    & \textbf{64.97}    & \underline{65.35} & \underline{63.20} & \underline{50.02} & \underline{63.09}                \\
    \textbf{\textsc{LACA\textsubscript{Orca\textsubscript{13}}}}  & \underline{71.61} & \underline{64.25} & \textbf{65.41}    & \textbf{63.46}    & \textbf{51.15}    & \textbf{63.18}                   \\
    \textbf{\textsc{LACA\textsubscript{LLaMA\textsubscript{8}}}}  & 71.17             & 63.81             & 64.29             & 61.46             & 49.71             & 62.09                            \\
    \textbf{\textsc{LACA\textsubscript{LLaMA\textsubscript{3}}}}  & 69.19             & 62.32             & 62.23             & 58.93             & 47.02             & 59.94                            \\
    \textbf{\textsc{LACA\textsubscript{LLaMA\textsubscript{1}}}}  & 69.00             & 60.72             & 61.79             & 56.79             & 46.61             & 58.98                            \\ \bottomrule
    \end{tabular}
    \end{adjustbox}
    \caption{Results with different LLMs for generation with English as the source language and other languages as the target ones compared to \textsc{Zero-shot} results. The best result for each model and target language is in \textbf{bold}; the second best is \underline{underlined}.}
    \label{tab:res_small}
\end{table}

For instance, relative to \textsc{LACA} with the 8B LLaMA~3.1 model, \textbf{\textsc{LACA\textsubscript{LLaMA\textsubscript{3}}}} shows an average performance drop of approximately 4\% for mBERT and approximately 1\% for XLM-R. \textbf{\textsc{LACA\textsubscript{LLaMA\textsubscript{1}}}} performs about 6\% worse for mBERT and 3\% worse for XLM-R. The largest declines occur for Russian and Turkish – languages not officially supported by LLaMA or closely related to those supported – highlighting the importance of LLM size for underrepresented languages.

Interestingly, the gap between the 8B and 70B LLaMA models is smaller than the gap between the 3B and 8B models, suggesting diminishing returns with larger sizes. For resource-constrained scenarios, smaller models remain a viable option, though their limitations should be considered. Balancing size, computational requirements, and language coverage is crucial for optimal performance when selecting an LLM.

\begin{table*}[ht!]
\centering
\begin{adjustbox}{width=\linewidth}
\begin{tabular}{@{}clcccccccccccc@{}}
\toprule
\multirow{2}{*}{\textbf{\begin{tabular}[c]{@{}c@{}}Source\\ Language\end{tabular}}} & \multirow{2}{*}{\textbf{Method}}                            & \multicolumn{5}{c}{\textbf{mBERT}}                                                 &                & \multicolumn{5}{c}{\textbf{XLM-R}}                                                 &                \\ \cmidrule(lr){3-8} \cmidrule(lr){9-14}
&                                                             & En             & Es             & Fr             & Nl             & Ru             & Tr             & En             & Es             & Fr             & Nl             & Ru             & Tr             \\ \midrule
& \textsc{Supervised}                                           & 65.39          & 67.88          & 61.80          & 56.80          & 58.87          & 47.74          & 73.81          & 71.93          & 67.44          & 64.28          & 64.93          & 60.93          \\ \cdashlinelr{1-14}
\multirow{4}{*}{En}                                                                 & \textsc{Zero-shot}                                            & --              & 56.90          & 45.80          & 45.97          & 34.06          & 27.04          & --              & 67.48          & 58.87          & 58.95          & 56.10          & 46.53          \\\cdashlinelr{2-14}
        & \textbf{\textsc{LACA\textsubscript{LLaMA\textsubscript{70}}}} & --              & \textbf{65.23} & \textbf{54.90} & \underline {55.29}    & \underline {53.72}    & \underline {31.98}    & --              & 71.89          & \textbf{64.97} & 65.35          & \underline {63.20}    & \underline {50.02}    \\
        & \textbf{\textsc{LACA\textsubscript{Orca\textsubscript{13}}}}  & --              & \underline {64.80}    & \underline {54.21}    & \textbf{55.41} & \textbf{53.86} & \textbf{33.16} & --              & 71.61          & \underline {64.25}    & 65.41          & \textbf{63.46} & \textbf{51.15} \\
        & \textbf{\textsc{LACA\textsubscript{LLaMA\textsubscript{8}}}}  & --              & 64.33          & 53.74          & 54.56          & 52.36          & 31.13          & --              & 71.17          & 63.81          & 64.29          & 61.46          & 49.71          \\\cdashlinelr{1-14}
\multirow{4}{*}{Es}                                                                 & \textsc{Zero-shot}                                            & 45.76          & --              & 42.70          & 38.29          & 25.25          & 16.24          & 56.86          & --              & 56.10          & 59.41          & 56.43          & 41.20          \\\cdashlinelr{2-14}
        & \textbf{\textsc{LACA\textsubscript{LLaMA\textsubscript{70}}}} & \underline {56.12}    & --              & 53.79          & 48.25          & 41.28          & 21.15          & 69.71          & --              & 59.99          & 63.52          & 58.00          & 45.31          \\
        & \textbf{\textsc{LACA\textsubscript{Orca\textsubscript{13}}}}  & \textbf{57.45} & --              & 51.71          & 52.17          & 40.52          & 23.29          & 68.77          & --              & 60.46          & \underline {65.47}    & 58.36          & 45.47          \\
        & \textbf{\textsc{LACA\textsubscript{LLaMA\textsubscript{8}}}}  & 55.84          & --              & 50.83          & 47.20          & 40.52          & 21.54          & 70.17          & --              & 59.73          & 62.57          & 57.29          & 45.08          \\\cdashlinelr{1-14}
\multirow{4}{*}{Fr}                                                                 & \textsc{Zero-shot}                                            & 44.10          & 54.46          & --              & 37.79          & 29.95          & 19.42          & 51.62          & 67.43          & --              & 59.60          & 52.82          & 36.60          \\\cdashlinelr{2-14}
        & \textbf{\textsc{LACA\textsubscript{LLaMA\textsubscript{70}}}} & 53.82          & 63.32          & --             & 49.32          & 37.85          & 22.32          & 59.14          & \underline {73.09}    & --              & 63.71          & 55.60          & 39.85          \\
        & \textbf{\textsc{LACA\textsubscript{Orca\textsubscript{13}}}}  & 53.88          & 63.97          & --              & 50.11          & 38.89          & 22.38          & 57.82          & \textbf{74.54} & --              & 64.62          & 56.43          & 40.02          \\
        & \textbf{\textsc{LACA\textsubscript{LLaMA\textsubscript{8}}}}  & 52.96          & 63.49          & --              & 49.11          & 38.27          & 21.78          & 57.50          & 72.99          & --              & 64.62          & 55.27          & 39.41          \\\cdashlinelr{1-14}
\multirow{4}{*}{Nl}                                                                 & \textsc{Zero-shot}                                            & 45.68          & 45.53          & 36.20          & --              & 27.62          & 28.32          & 62.30          & 65.69          & 54.43          & --              & 56.19          & 43.90          \\\cdashlinelr{2-14}
        & \textbf{\textsc{LACA\textsubscript{LLaMA\textsubscript{70}}}} & 53.77          & 57.83          & 47.81          & --              & 38.52          & 31.44          & 66.07          & 70.16          & 62.04          & --              & 59.00          & 48.31          \\
        & \textbf{\textsc{LACA\textsubscript{Orca\textsubscript{13}}}}  & 53.06          & 58.72          & 47.56          & --              & 39.10          & 31.28          & 67.39          & 69.49          & 63.62          & --              & 60.62          & 48.27          \\
        & \textbf{\textsc{LACA\textsubscript{LLaMA\textsubscript{8}}}}  & 53.36          & 58.14          & 48.16          & --              & 38.03          & 31.00          & 66.73          & 68.07          & 60.42          & --              & 58.58          & 47.74          \\\cdashlinelr{1-14}
\multirow{4}{*}{Ru}                                                                 & \textsc{Zero-shot}                                            & 37.42          & 49.62          & 33.00          & 35.77          & --              & 24.24          & 65.09          & 63.20          & 57.60          & 59.39          & --              & 44.62          \\\cdashlinelr{2-14}
        & \textbf{\textsc{LACA\textsubscript{LLaMA\textsubscript{70}}}} & 53.36          & 57.43          & 44.25          & 46.29          & --              & 29.14          & \underline {70.45}    & 69.02          & 62.71          & \textbf{65.53} & --              & 48.99          \\
        & \textbf{\textsc{LACA\textsubscript{Orca\textsubscript{13}}}}  & 55.75          & 57.18          & 43.06          & 46.68          & --              & 29.57          & \textbf{71.18} & 67.41          & 63.39          & 65.17          & --              & 49.21          \\
        & \textbf{\textsc{LACA\textsubscript{LLaMA\textsubscript{8}}}}  & 52.65          & 58.12          & 41.63          & 45.25          & --              & 28.97          & 70.25          & 68.93          & 61.79          & 65.12          & --              & 49.02          \\\cdashlinelr{1-14}
\multirow{4}{*}{Tr}                                                                 & \textsc{Zero-shot}                                            & 38.67          & 41.54          & 29.01          & 28.53          & 34.19          & --              & 56.33          & 49.84          & 48.08          & 49.90          & 46.26          & --              \\\cdashlinelr{2-14}
        & \textbf{\textsc{LACA\textsubscript{LLaMA\textsubscript{70}}}} & 49.42          & 51.94          & 40.77          & 40.47          & 40.98          & --              & 62.45          & 54.17          & 54.94          & 54.58          & 53.26          & --              \\
        & \textbf{\textsc{LACA\textsubscript{Orca\textsubscript{13}}}}  & 49.11          & 50.02          & 39.64          & 41.08          & 41.29          & --              & 61.43          & 53.82          & 55.34          & 53.50          & 53.47          & --              \\
        & \textbf{\textsc{LACA\textsubscript{LLaMA\textsubscript{8}}}}  & 49.20          & 49.53          & 39.81          & 40.92          & 40.59          & --              & 60.34          & 54.50          & 53.89          & 52.92          & 52.89          & --             \\ \bottomrule
\end{tabular}
\end{adjustbox}
\caption{Results for different combinations of source and target languages. The best result for each target language is in \textbf{bold}; the second best is \underline{underlined}.}
\label{tab:res_all}
\end{table*}

\subsection{Results by Source-Target Language Combinations}

Table~\ref{tab:res_all} presents the results for different combinations of source and target languages, highlighting the effectiveness of the proposed \textsc{LACA} approach. Across all language combinations, \textsc{LACA} significantly improves upon the \textsc{Zero-shot} results, demonstrating its robust performance. The results confirm the strong potential of all three employed LLMs for generating pseudo-labelled data, further enhancing the cross-lingual performance.

The results reveal that English is the most effective source language in most cases. Additionally, selecting source languages from the same language branch appears advantageous. For example, the combination of French and Spanish, both Romance (Latin) languages, yields excellent results. Conversely, Turkish performs the worst, both as a source and a target language. This poor performance could be attributed to Turkish's status as the only language from the Turkic family, making it distinct from the Indo-European languages, and its limited proximity to the languages officially supported by LLaMA~3.1.

Interestingly, especially for XLM-R, Russian is an effective source language despite its use of the Cyrillic alphabet, in contrast to the Latin alphabet used by most other languages. Furthermore, Russian does not belong to the same branch of the Indo-European language family as any of the other languages used. We speculate that this effectiveness might stem from the larger size of its training dataset compared to other languages, as even \textsc{Zero-shot} results are often better with Russian as the source language.

\section{Error Analysis}
\label{sec:error_analysis}

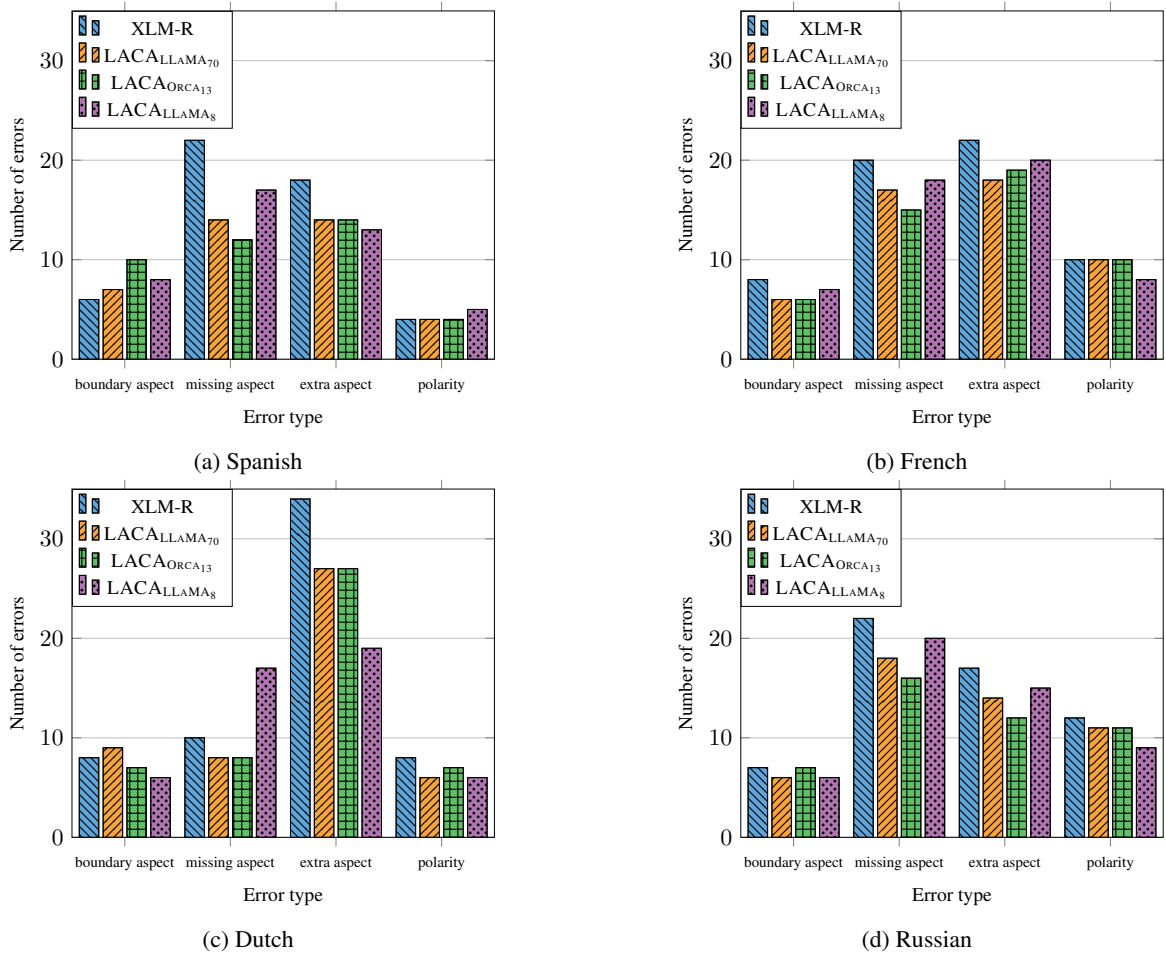
\begin{figure*}[ht!]
    \begin{subfigure}{0.45\textwidth}
        \centering
        \begin{adjustbox}{width=0.9\linewidth}
            \begin{tikzpicture}
                \begin{axis}[
                    ybar,
                    bar width=9pt,
                    xtick={0,1,2,3},
                    ytick={0,10,20,30},
                    xticklabels={\scriptsize \strut boundary aspect,\scriptsize \strut missing aspect,\scriptsize \strut extra aspect,\scriptsize \strut polarity},
                    ymin=0,
                    ymax=35,
                    xmin=-0.5,
                    xmax=3.5,
                    ymajorgrids=true,
                    ylabel={\footnotesize Number of errors},
                    xlabel={\footnotesize Error type},
                    legend style={at={(0,1)}, anchor=north west, font=\footnotesize},
                    ]
                    \addplot[black,fill=lightblue,postaction={pattern=north west lines}] coordinates {(0,6) (1,22) (2,18) (3,4)};
                    \addplot[black,fill=lightorange,postaction={pattern=north east lines}] coordinates {(0,7) (1,14) (2,14) (3,4)};
                    \addplot[black,fill=lightgreen,postaction={pattern=grid}] coordinates {(0,10) (1,12) (2,14) (3,4)};
                    \addplot[black,fill=lightpurple,postaction={pattern=crosshatch dots}] coordinates {(0,8) (1,17) (2,13) (3,5)};
                    \legend{XLM-R,\textsc{LACA\textsubscript{LLaMA\textsubscript{70}}}, \textsc{LACA\textsubscript{Orca\textsubscript{13}}}, \textsc{LACA\textsubscript{LLaMA\textsubscript{8}}}}
                \end{axis}
            \end{tikzpicture}
        \end{adjustbox}
       \subcaption{Spanish}
    \end{subfigure}
    \hfill
    \begin{subfigure}{0.45\textwidth}
        \centering
        \begin{adjustbox}{width=0.9\linewidth}
            \begin{tikzpicture}
                \begin{axis}[
                    ybar,
                    bar width=9pt,
                    xtick={0,1,2,3},
                    ytick={0,10,20,30},
                    xticklabels={\scriptsize \strut boundary aspect,\scriptsize \strut missing aspect,\scriptsize \strut extra aspect,\scriptsize \strut polarity},
                    ymin=0,
                    ymax=35,
                    xmin=-0.5,
                    xmax=3.5,
                    ymajorgrids=true,
                    ylabel={\footnotesize Number of errors},
                    xlabel={\footnotesize Error type},
                    legend style={at={(0,1)}, anchor=north west, font=\footnotesize},
                    ]
                    \addplot[black,fill=lightblue,postaction={pattern=north west lines}] coordinates {(0,8) (1,20) (2,22) (3,10)};
                    \addplot[black,fill=lightorange,postaction={pattern=north east lines}] coordinates {(0,6) (1,17) (2,18) (3,10)};
                    \addplot[black,fill=lightgreen,postaction={pattern=grid}] coordinates {(0,6) (1,15) (2,19) (3,10)};
                    \addplot[black,fill=lightpurple,postaction={pattern=crosshatch dots}] coordinates {(0,7) (1,18) (2,20) (3,8)};
                    \legend{XLM-R,\textsc{LACA\textsubscript{LLaMA\textsubscript{70}}}, \textsc{LACA\textsubscript{Orca\textsubscript{13}}}, \textsc{LACA\textsubscript{LLaMA\textsubscript{8}}}}
                \end{axis}
            \end{tikzpicture}
        \end{adjustbox}
       \subcaption{French}
    \end{subfigure}
    \hfill
    \begin{subfigure}{0.45\textwidth}
        \centering
        \begin{adjustbox}{width=0.9\linewidth}
            \begin{tikzpicture}
                \begin{axis}[
                    ybar,
                    bar width=9pt,
                    xtick={0,1,2,3},
                    ytick={0,10,20,30},
                    xticklabels={\scriptsize \strut boundary aspect,\scriptsize \strut missing aspect,\scriptsize \strut extra aspect,\scriptsize \strut polarity},
                    ymin=0,
                    ymax=35,
                    xmin=-0.5,
                    xmax=3.5,
                    ymajorgrids=true,
                    ylabel={\footnotesize Number of errors},
                    xlabel={\footnotesize Error type},
                    legend style={at={(0,1)}, anchor=north west, font=\footnotesize},
                    ]
                    \addplot[black,fill=lightblue,postaction={pattern=north west lines}] coordinates {(0,8) (1,10) (2,34) (3,8)};
                    \addplot[black,fill=lightorange,postaction={pattern=north east lines}] coordinates {(0,9) (1,8) (2,27) (3,6)};
                    \addplot[black,fill=lightgreen,postaction={pattern=grid}] coordinates {(0,7) (1,8) (2,27) (3,7)};
                    \addplot[black,fill=lightpurple,postaction={pattern=crosshatch dots}] coordinates {(0,6) (1,17) (2,19) (3,6)};
                    \legend{XLM-R,\textsc{LACA\textsubscript{LLaMA\textsubscript{70}}}, \textsc{LACA\textsubscript{Orca\textsubscript{13}}}, \textsc{LACA\textsubscript{LLaMA\textsubscript{8}}}}
                \end{axis}
            \end{tikzpicture}
        \end{adjustbox}
       \subcaption{Dutch}
    \end{subfigure}
    \hfill
    \begin{subfigure}{0.45\textwidth}
        \centering
        \begin{adjustbox}{width=0.9\linewidth}
            \begin{tikzpicture}
                \begin{axis}[
                    ybar,
                    bar width=9pt,
                    xtick={0,1,2,3},
                    ytick={0,10,20,30},
                    xticklabels={\scriptsize \strut boundary aspect,\scriptsize \strut missing aspect,\scriptsize \strut extra aspect,\scriptsize \strut polarity},
                    ymin=0,
                    ymax=35,
                    xmin=-0.5,
                    xmax=3.5,
                    ymajorgrids=true,
                    ylabel={\footnotesize Number of errors},
                    xlabel={\footnotesize Error type},
                    legend style={at={(0,1)}, anchor=north west, font=\footnotesize},
                    ]
                    \addplot[black,fill=lightblue,postaction={pattern=north west lines}] coordinates {(0,7) (1,22) (2,17) (3,12)};
                    \addplot[black,fill=lightorange,postaction={pattern=north east lines}] coordinates {(0,6) (1,18) (2,14) (3,11)};
                    \addplot[black,fill=lightgreen,postaction={pattern=grid}] coordinates {(0,7) (1,16) (2,12) (3,11)};
                    \addplot[black,fill=lightpurple,postaction={pattern=crosshatch dots}] coordinates {(0,6) (1,20) (2,15) (3,9)};
                    \legend{XLM-R,\textsc{LACA\textsubscript{LLaMA\textsubscript{70}}}, \textsc{LACA\textsubscript{Orca\textsubscript{13}}}, \textsc{LACA\textsubscript{LLaMA\textsubscript{8}}}}
                \end{axis}
            \end{tikzpicture}
        \end{adjustbox}
       \subcaption{Russian}
    \end{subfigure}
    \caption{Number of error types in 100 samples for different languages with English as the source language and XLM-R as the backbone model.}
    \label{fig:error}
\end{figure*}

We performed a detailed error analysis to gain insight into the most common errors made by the models. We manually examined 100 samples from the test sets of four languages using the best-performing model trained on English, with XLM-R as the backbone. The analysis focused on four main types of errors:
\begin{enumerate}
    \item \textbf{Boundary aspect errors:} These occur when the model either misses part of an aspect term or includes extra words.
    \item \textbf{Missing aspects:} These errors arise when the model entirely fails to detect an aspect term present in the gold labels.
    \item \textbf{Extra aspects:} These occur when the model predicts an aspect term that is not present in the gold labels.
    \item \textbf{Sentiment polarity errors:} These involve incorrect sentiment classification for correctly identified aspects or boundary aspect errors.
\end{enumerate}

The results, shown in Figure~\ref{fig:error}, indicate that boundary aspect errors and sentiment polarity errors are relatively less frequent. Interestingly, error distribution varies across languages. For instance, extra aspect errors are significantly more common in Dutch than other error types. In contrast, for other languages, errors are more balanced. These differences may be influenced by the distribution of labels in the datasets; in the Dutch test set, there are fewer aspects per sentence than in other languages.

The proposed \textsc{LACA} framework reduces the total number of errors by decreasing missing and extra aspect errors. However, for Spanish, it slightly increased boundary aspect errors. This minor increase may not be entirely negative, as boundary aspect errors often indicate predictions closer to the gold labels than missing or extra aspect errors.

An interesting observation was made for \textsc{LACA\textsubscript{LLaMA\textsubscript{8}}} in Dutch. This model notably increased the number of missing aspect errors but significantly reduced extra aspect errors, highlighting a trade-off in its error patterns.

\end{document}